\documentclass[letterpaper]{article}
\usepackage[preprint]{aaai2027}
\usepackage[hyphens]{url}
\usepackage{graphicx}
\usepackage{natbib}
\usepackage{caption}
\usepackage{amsmath,amsfonts,amssymb}
\usepackage{algorithm}
\usepackage{algorithmic}
\usepackage{array}
\usepackage{booktabs}
\usepackage{makecell}
\usepackage{multirow}
\usepackage{textcomp}
\usepackage{xspace}
\usepackage{pifont}

\newcommand{\ours}{ThinkAfford}

\newcommand{\best}[1]{\textbf{#1}}

\title{ThinkAfford: Affordance-Centric Reasoning for Fine-Grained 3D Grounding in Cluttered Scenes}
\author{
    Xinrui Lin\textsuperscript{\rm 1,\textdagger},
    Sha Zhang\textsuperscript{\rm 2,\textdagger},
    Shumin Wang\textsuperscript{\rm 1},\\
    Zenghuan Zhu\textsuperscript{\rm 1},
    Jiajun Deng\textsuperscript{\rm 1,*},
    Yanyong Zhang\textsuperscript{\rm 1,*}
}
\affiliations{
    \textsuperscript{\rm 1}University of Science and Technology of China, Hefei, China\\
    \textsuperscript{\rm 2}The Chinese University of Hong Kong, Hong Kong SAR, China
}

\begin{document}

\maketitle
\begingroup
\renewcommand{\thefootnote}{\fnsymbol{footnote}}
\footnotetext[2]{Equal contribution.}
\footnotetext[1]{Correspondence: dengjj@ustc.edu.cn; yanyongz@ustc.edu.cn.}
\endgroup
\begin{abstract}
Task-driven 3D affordance grounding aims to localize the functional region in a cluttered 3D scene that enables an action specified by a natural-language instruction. 
Existing methods either predict 3D masks directly or construct them by selecting and fusing intermediate 2D/3D regions. However, they remain vulnerable to two intertwined failure modes: the predicted or selected regions may miss the target interaction area or have unsuitable granularity, while language grounding may confuse visually similar alternatives under relational instructions.
To this end, We introduce \ours{}, which decouples high-recall affordance proposal generation from instruction-grounded reasoning. 
Specifically, the Affordance Proposal Generation module first uses learnable affordance prompts and multi-level visual features to predict interaction-conditioned heatmaps, extracting a variable number of fine-grained proposals without parsed object or part names as segmentation prompts.
Visual-Prompted Affordance Reasoning then reasons over labeled proposal overlays using the full instruction, returning identifiers in a structured ``think-then-answer'' response.
Moreover, Group Relative Policy Optimization uses proposal-level rewards from lifted 3D overlap to align VPAR selection with final 3D grounding.
On the SceneFun3D validation split, \ours{} achieves 10.69\% AP$_{50}$ and 25.46\% AP$_{25}$ under the official evaluator, outperforming comparable 3D open-vocabulary and vision-language-model-based 2D-to-3D baselines.
Module-level diagnostics further show that APG attains 77.5\% recall at 25\% intersection-over-union, while GRPO-trained VPAR achieves 72.1\% selection accuracy on APG-covered queries, compared with 63.4\% under supervised fine-tuning. 
\end{abstract}

\section{1 Introduction}

Grounding a task instruction to a functional part in a 3D environment is a fundamental capability for embodied agents. 
Given a reconstructed indoor scene and a natural-language instruction, an agent must find the 3D region that enables the requested interaction.
Unlike conventional object grounding, the target is often a small contact region rather than an entire object. For example, ``open the top-right drawer of the cabinet with the TV on top'' requires the agent to find the correct cabinet, identify the top-right drawer, and finally localize its pullable region. Predicting the entire cabinet or drawer is insufficient for interaction.
Such targets are often small, occluded, and similar to nearby parts, making them difficult to localize in cluttered scenes~\cite{wu2025segment,zheng2025densegrounding,li2024manipllm}.

Direct methods ground language in point clouds, radiance fields, or 3D instance masks~\cite{takmaz2023openmask3d,kerr2023lerf,li2024laso,chu20253d}. Although these representations preserve geometric consistency across views, existing 3D features often lack the spatial granularity needed to delineate small functional regions, while language-conditioned 3D models remain limited in resolving compositional and relational instructions.

These limitations have motivated 2D vision-language reasoning in 3D grounding.
Fun3DU~\cite{corsetti2025functionality} parses the instruction into functional and contextual objects, asks a VLM to place point prompts in selected views, converts them into masks with a promptable segmentor, and lifts the masks into 3D.
TASA~\cite{he2026task} builds on this point-to-mask route with task-aware view processing and geometric refinement, while Scene-R1~\cite{yuan2025scene} trains temporal and image grounding with GRPO before 2D segmentation and lifting.
AffordBot~\cite{wang2026affordbot} instead extracts 3D affordance elements with Mask3D~\cite{schult2023mask3d}, projects their identifiers and geometry-semantic descriptors into surround views, and prompts an MLLM to select them.

Whether produced in 2D or 3D, these intermediate regions define the candidates for later selection or fusion; we call each a \emph{proposal}. This common structure exposes two problems.
First, the proposal set may omit the interaction region or represent it at the wrong granularity. With point-to-mask methods, a parsed concept such as ``drawer'' can lead a promptable segmentor to return the drawer body rather than its handle or contact region; with 3D candidates such as those used by AffordBot, the small region may be absent from the Mask3D proposal set.
Second, selecting among repeated, functionally similar proposals requires jointly satisfying spatial, ordinal, and part--whole constraints that prompting alone may not reliably resolve.
Consequently, stronger language reasoning cannot recover a missing target, and high proposal recall alone does not ensure correct selection.

We therefore formulate task-driven 3D affordance grounding as two complementary problems: \emph{high-recall affordance proposal generation}, which ensures that the functional region is represented, and \emph{instruction-grounded affordance reasoning}, which selects it.
Based on this decomposition, we present \ours{}, with APG and VPAR addressing the two problems, respectively.
\textbf{Affordance Proposal Generation (APG)} predicts affordance-conditioned heatmaps instead of prompting a generic segmentor with parsed object or part names.
Learnable affordance context and multi-level features preserve functional semantics and local detail; variable-number extraction retains small handles, switches, buttons, and other contact regions rather than one object- or part-level mask.
\textbf{Visual-Prompted Affordance Reasoning (VPAR)} selects identifiers from a labeled APG overlay using the image and full instruction.
Its structured think-then-answer format makes relation-aware selection verifiable.
We train VPAR with Group Relative Policy Optimization (GRPO)~\cite{shao2024deepseekmath}, rewarding sampled choices by lifted 3D overlap with the ground truth.
Within-group reward differences compare scene proposals to teach spatial, ordinal, and part--whole relations.

Experiments on SceneFun3D show that \ours{} outperforms comparable baselines under the official evaluation protocol, while module-level analyses validate both APG proposal coverage and GRPO-based selection.

Our contributions are summarized as follows:

\begin{itemize}

\item We decompose task-driven 3D affordance grounding into high-recall proposal generation and instruction-grounded reasoning, enabling separate measurement of proposal misses and selection errors.

\item We develop APG to produce interaction-level proposals rather than object- or part-level masks. It combines affordance-specific context learning, multi-level image features, and variable-number proposal extraction to retain small functional regions with high recall.

\item We propose VPAR, a structured think-then-answer policy that selects labeled proposals under the full instruction. GRPO optimization with rewards computed from lifted 3D overlap improves relation-sensitive selection over supervised fine-tuning.
\end{itemize}

\section{2 Related Work}

\paragraph{3D Affordance Grounding.}
Earlier image-based methods learn affordance maps from 2D supervision or demonstrations, but typically operate on object-centric views without scene-level 3D masks. SceneFun3D~\cite{delitzas2024scenefun3d} introduces task-driven language and 3D functional-part annotations for cluttered scenes. OpenMask3D~\cite{takmaz2023openmask3d}, LERF~\cite{kerr2023lerf}, and OpenMask3D-F~\cite{delitzas2024scenefun3d} support open-vocabulary or functional 3D grounding, but their object-scale point-cloud or feature representations can miss thin interaction regions specified by actions. Preserving geometry across views therefore does not guarantee localization at the interaction granularity.

\paragraph{VLM-Based 2D-to-3D Grounding.}
Fun3DU~\cite{corsetti2025functionality} parses functional and contextual objects, converts VLM point prompts into masks, and aggregates them in 3D; TASA~\cite{he2026task} adds affordance-weighted frame selection, point validation, and geometric refinement. Scene-R1~\cite{yuan2025scene} trains temporal and image grounding with GRPO before box prediction and segmentation, whereas AffordBot~\cite{wang2026affordbot} asks an MLLM to select projected Mask3D affordance elements and infer motion. Despite these different pipelines, all rely on intermediate candidates, so errors in object hypotheses, localization, or granularity propagate to the final 3D mask.

\subsection{Reinforcement Learning for Visual Reasoning}
GRPO~\cite{shao2024deepseekmath} learns from groups of sampled responses and has been used to improve reasoning models. Recent visual reasoning systems use reinforcement learning to improve view selection, answer format, or chain-of-thought behavior~\cite{yuan2025scene,wang2025affordance}. In particular, Scene-R1~\cite{yuan2025scene} shows that GRPO can improve video-level snippet selection and downstream box grounding. In our work, the policy action is instead the selection of affordance-proposal identifiers. Proposal targets are constructed by lifting each candidate to the annotated 3D scene, so reward differences correspond to concrete choices among physically distinct functional regions.

\begin{figure*}[!htbp]
\centering
\includegraphics[width=\textwidth]{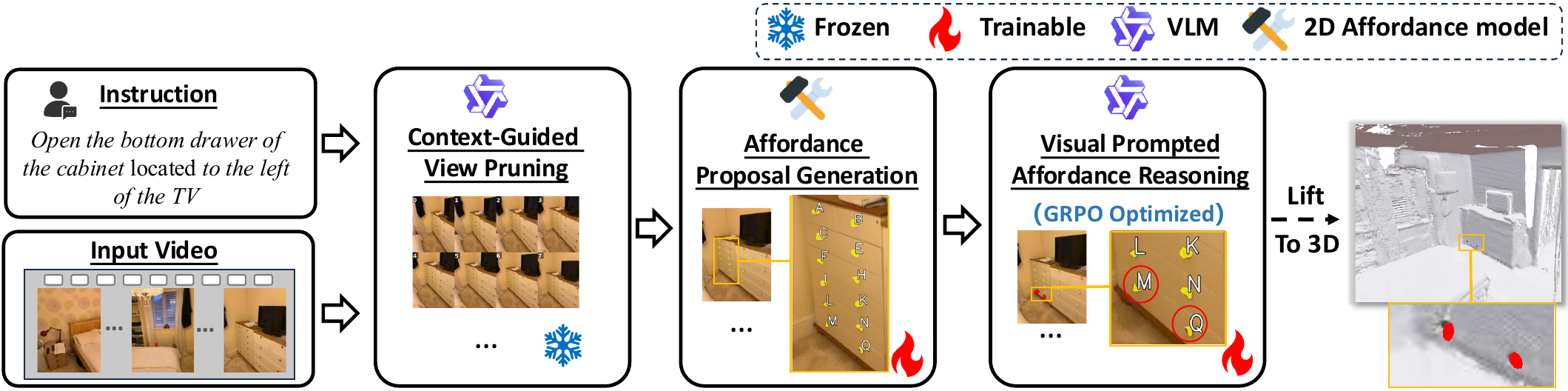}
\caption{Overview of \ours. The method keeps a compact set of views, generates affordance-centric proposals, selects proposal identifiers with a GRPO-trained VLM, and fuses the selected evidence into a 3D mask.}
\label{fig:framework}
\end{figure*}

\section{3 Method}
Let $\mathcal{P}=\{p_i\}_{i=1}^{N}$ be a 3D scene point cloud and 
let $\mathcal{I}=\{I_v\}_{v\in\mathcal{V}}$ be  RGB views with known 
camera parameters. 
Given a natural-language instruction $q$, the goal is 
to predict a 3D affordance mask $\mathcal{M}^{3D}\subseteq\mathcal{P}$ for the 
interactable region described by $q$.

As shown in Fig.~\ref{fig:framework}, \ours{} first uses a VLM to parse the instruction and retain video views in which the contextual object is visible. APG then generates affordance proposals for each retained view, and VPAR selects the proposal(s) that best match the instruction. The selected 2D masks are lifted and fused into the final 3D affordance mask.
This design makes proposal coverage and instruction-grounded reasoning separately measurable.

\subsection{3.1 Context-Guided View Pruning}
We prune the video into a compact view set $\mathcal{V}^{\star}$, aiming to reduce computation and eliminate views irrelevant to the instructed region before finer-grained reasoning.

Given an instruction $q$, we first prompt a VLM to extract the target object or part, contextual anchors, and coarse relations. 
For example, in ``open the top-left drawer of the cabinet with the TV on its top,'' the cabinet is used as the view-pruning anchor, while the TV provides scene-level context for disambiguating the correct cabinet. 
When no part--whole relation exists, the target itself serves as the anchor. 

We uniformly sample the input video at a fixed interval and query the VLM with a binary visibility prompt to identify where the anchor is visible. 
Around each positive frame, we retain a fixed temporal neighborhood to include nearby viewpoints of the same object and to reduce sensitivity to a single VLM decision. 
When the instruction contains a coarse contextual relation, we issue a second verification prompt, such as whether the TV is on top of the cabinet, and discard frames that are inconsistent with the parsed relation. 
The retained frames form $\mathcal{V}^{\star}$, providing compact contextual evidence for subsequent fine-grained grounding.

\subsection{3.2 Affordance Proposal Generation}
\label{subsec:apg}
Generic open-vocabulary segmentors produce object- or named-part masks, which may be too coarse for interaction. For example, ``drawer'' can yield the drawer front rather than its pullable handle. 
We therefore introduce APG to generate a high-recall set of interaction-level candidates.

Given $q$, the parsing VLM predicts an affordance type $a^\star \in \mathcal{A}$, such as \texttt{foot\_push} or \texttt{pinch\_pull}. From $(I_v,a^\star)$, APG predicts a heatmap $S_v^{a^\star}$ and extracts a proposal set $\mathcal{M}_v = \{M_{v,j}\}_{j=1}^{J_v}$. This interaction conditioning avoids exact object or part names as segmentation prompts; VPAR later uses the full instruction to select candidates.

\begin{figure}[!htbp]
\centering
\includegraphics[width=\linewidth]{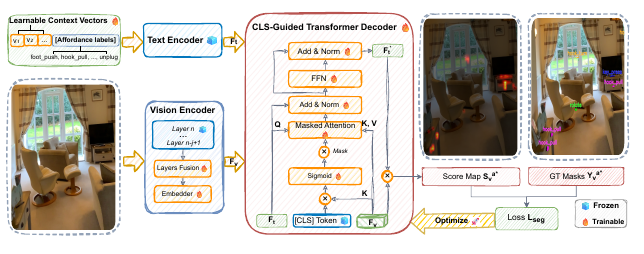}
    \caption{An illustration of the APG module. Frozen CLIP/DINOv2 encoders with learnable context vectors feed a CLS-guided decoder to produce affordance heatmaps.}

\label{fig:apg-architecture}
\end{figure}

\paragraph{Data Preparation from 3D Labels.}
\label{subsubsec:apg_training}
APG requires 2D affordance masks, whereas the dataset provides labels only on the 3D point cloud. 
Direct projection causes occlusion bleed-through, because hidden annotated points may appear in the 2D images.
To address this, we render the point cloud for each RGB view and build a visibility-aware index map $\mathrm{Id}_v$. A depth-tolerant z-buffer resolves most pixels; ambiguous pixels are resolved by fitting a local plane and selecting the closest ray--plane intersection. Thus, only visible points contribute to the target.
For each affordance type $a \in \mathcal{A}$, we project its 3D instance masks $\{\mathcal{M}^{3D}_{a,k}\}_{k=1}^{K_a}$ and filter them with $\mathrm{Id}_v$. Their visible union forms the 2D target $Y_v^a$, providing visibility-consistent supervision.
The geometry-derived $Y_v^a$ is used only as supervision. 

\paragraph{Architecture and Training.}
As shown in Fig.~\ref{fig:apg-architecture}, APG follows the OOAL architecture~\cite{li2024one}: frozen DINOv2~\cite{oquab2023dinov2} visual and CLIP~\cite{radford2021learning} text encoders with a lightweight CLS-guided cross-attention decoder.
To handle small affordance regions in cluttered scenes, we use CoOp~\cite{zhou2022learning} for affordance-specific prompts and fuse multi-level features to preserve spatial detail. These adaptations make the heatmap localize interaction regions rather than the full named object.

During training, APG takes $I_v$ and the ground-truth type $a$ to predict $S_v^a$, supervised by the visibility-aware target $Y_v^a$ through pixel-wise binary cross-entropy:

\begin{equation}
\label{eq:apg_loss}
\mathcal{L}_{\mathrm{seg}} = \mathrm{BCE} \left( S_v^a, Y_v^a \right).
\end{equation}
The encoders remain frozen; only the context vectors and decoder are updated. At inference, $a$ is replaced by the VLM-predicted type $a^\star$.

\subsubsection{Proposal Extraction}

The resulting heatmap $S_v^{a^\star}$ is resized to the original image resolution and thresholded at $\gamma$. DBSCAN then separates the foreground into disconnected components without assuming a fixed instance count, yielding the variable-length proposal set $\mathcal{M}_v$. To maximize recall, APG retains all confident components, including repeated regions such as drawer handles, and assigns each a unique letter (A, B, \dots, Z, AA, \dots) for downstream selection.

\begin{figure*}[!t]
\centering
\includegraphics[width=\textwidth]{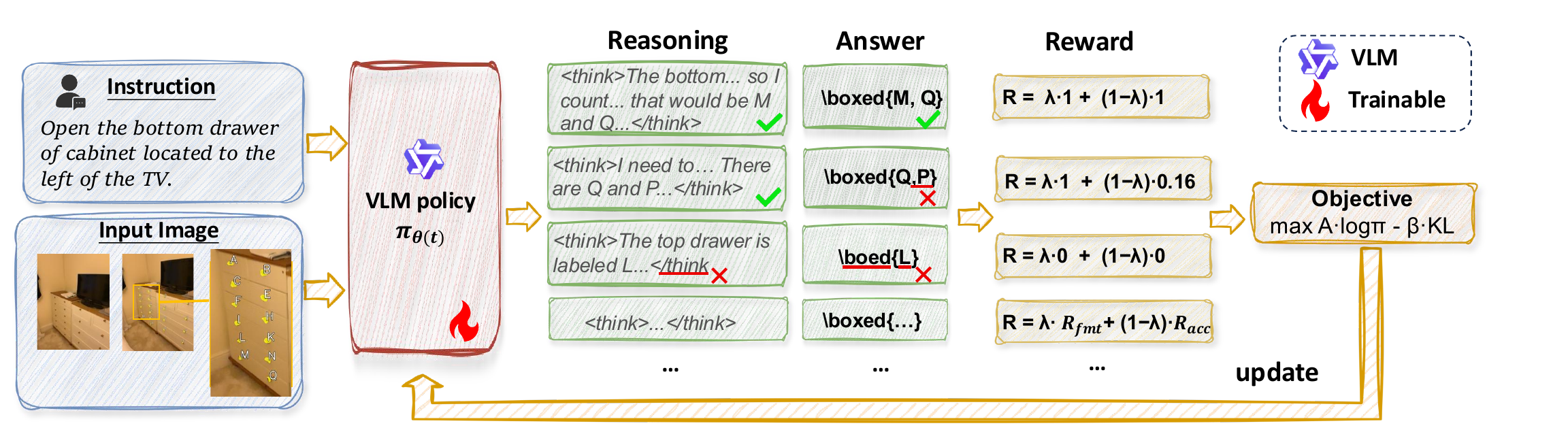}
\caption{GRPO training for VPAR. Given an instruction and a labeled proposal overlay, the trainable VLM samples a group of structured think-then-answer responses. Each response is parsed into proposal identifiers and scored by format correctness and proposal-level grounding accuracy, producing group-relative advantages for policy optimization.}
\label{fig:grpo_training}
\end{figure*}

\subsection{3.3 Visual-Prompted Affordance Reasoning}
\label{subsec:vpar}
APG may return several regions with the same affordance type, such as handles on multiple drawers. 
Distinguishing them requires spatial, ordinal, or part--whole constraints. 
VPAR selects the proposal that satisfies the full instruction. 

\paragraph{Discrete Proposal Selection with Visual Prompts.}

For each selected view $I_v$, we overlay APG proposals and their identifiers to form $I'_v$. The original image supplies scene context, while the overlay defines a compact discrete action space. Given $(q,I_v,I'_v)$, VPAR outputs proposal identifiers, yielding
$\hat{\mathcal{C}}_v=\pi_\theta(q,I_v,I'_v)\subseteq\{1,\dots,J_v\}$.
Because the choice may depend on spatial, ordinal, or part--whole constraints, the prompt uses a structured think-then-answer format. The \texttt{\textless think\textgreater} field resolves the relevant object and relations, while \texttt{\textless answer\textgreater} returns identifiers, making the final choice verifiable. 

\paragraph{GRPO Optimization and Reward.}
As illustrated in Fig.~\ref{fig:grpo_training}, we train VPAR with Group Relative Policy Optimization (GRPO)~\cite{shao2024deepseekmath}. APG provides competing candidates, so sampled responses often select different physical regions. Their lifted 3D overlap yields relative rewards without manually constructed negatives, teaching the policy which choice better satisfies the instruction.

To construct targets, each proposal $M_{v,j}$ is lifted to $\mathcal{P}_{v,j}$ and compared with the ground-truth mask $\mathcal{M}^{3D}_{\mathrm{gt}}$ using point-level IoU,
$o_{v,j}=|\mathcal{P}_{v,j}\cap\mathcal{M}^{3D}_{\mathrm{gt}}|/|\mathcal{P}_{v,j}\cup\mathcal{M}^{3D}_{\mathrm{gt}}|$. Proposals with $o_{v,j}\ge\rho$ form the target set $\mathcal{C}_v^\star$. If none reaches $\rho$, the sample is an APG coverage failure and is excluded from VPAR training because no valid selection target exists; it remains in APG and end-to-end evaluation.

For each prompt, $\pi_\theta$ samples responses $\{y_i\}_{i=1}^{G}$, each parsed into $\hat{\mathcal{C}}_{v,i}$. The reward combines proposal accuracy and output-format correctness:

\begin{equation}
R(y_i,\mathcal{C}_v^\star) = 
(1-\lambda_{\mathrm{fmt}}) \cdot R_{\mathrm{acc}}(\hat{\mathcal{C}}_{v,i}, \mathcal{C}_v^\star)
+
\lambda_{\mathrm{fmt}} \cdot R_{\mathrm{fmt}}(y_i).
\end{equation}

$R_{\mathrm{fmt}}$ checks that the response is parsable and follows the required structure. We keep $\lambda_{\mathrm{fmt}}$ small so grounding accuracy dominates.

The accuracy reward is defined as

\begin{equation}
\begin{aligned}
R_{\mathrm{acc}}
&=
J(\hat{\mathcal{C}}_{v,i},\mathcal{C}_v^\star)
\left(
1-\frac{1}{2}
(r_{\mathrm{miss}}+r_{\mathrm{spur}})
\right),\\
r_{\mathrm{miss}}
&=
\frac{|\mathcal{C}_v^\star \setminus \hat{\mathcal{C}}_{v,i}|}{\max(1,|\mathcal{C}_v^\star|)},\\
r_{\mathrm{spur}}
&=
\frac{|\hat{\mathcal{C}}_{v,i} \setminus \mathcal{C}_v^\star|}{\max(1,|\hat{\mathcal{C}}_{v,i}|)}.
\end{aligned}
\end{equation}
where $J(\cdot,\cdot)$ is Jaccard similarity. The two penalties discourage missed and spurious selections, assigning low reward to a visually similar but incorrect neighbor.

Using the group mean $\bar{R}$, we compute the relative advantage

\begin{equation}
\hat{A}_i =
\frac{R(y_i)-\bar{R}}{\mathrm{std}(\{R(y_i)\}_{i=1}^{G})+\epsilon},
\end{equation}
and optimize the KL-regularized objective

\begin{equation}
\mathcal{J}_{\mathrm{GRPO}} = 
\mathbb{E}_{y_i \sim \pi_\theta}
\left[
\hat{A}_i \log \pi_\theta(y_i)
\right]
- 
\beta \cdot \mathrm{KL}
\left(
\pi_\theta \;\middle\|\; \pi_{\mathrm{ref}}
\right).
\end{equation}

Here, $\pi_{\mathrm{ref}}$ is the frozen reference policy and $\beta$ controls KL regularization.

\subsection{3.4 2D-to-3D Lifting and Fusion}
For each retained view, we merge the VPAR-selected proposals as $M_v^{\mathrm{sel}}=\bigcup_{j\in\hat{\mathcal{C}}_v}M_{v,j}$. We then re-query VPAR with the highlighted mask and use its yes-probability $w_v$ to down-weight unreliable views. Lifting $M_v^{\mathrm{sel}}$ with $\mathrm{Id}_v$ gives $\mathcal{P}_v\subseteq\mathcal{P}$. We fuse the multi-view evidence by weighted voting:
\begin{equation}
\sigma(p_i)=
\frac{\sum_{v\in\mathcal{V}^{\star}}w_v\mathbf{1}[p_i\in\mathcal{P}_v]}
{\sum_{v\in\mathcal{V}^{\star}}w_v}.
\end{equation}
The final mask is $\mathcal{M}^{3D}=\{p_i:\sigma(p_i)\ge\eta\}$. 

\section{4 Experiments}
\label{sec:experiments}

\subsection{4.1 Experimental Setup}

\paragraph{Datasets and metrics.}
We train on SceneFun3D's 200-scene training split and report in-domain results on its 30-scene validation split~\cite{delitzas2024scenefun3d}. To test distribution shift rather than only SceneFun3D appearance statistics, we additionally annotate 20 scenes with 15 instructions each from ScanNet~\cite{dai2017scannet}, ScanNet++~\cite{yeshwanth2023scannet++}, 3RScan~\cite{Wald2019RIO}, and MultiScan~\cite{mao2022multiscan}; these samples are used only for evaluation.
Following the official evaluator, we report AP$_{25}$ and AP$_{50}$ at IoU thresholds 0.25 and 0.50, and AP averaged from 0.50 to 0.95 in 0.05 increments when available. R@$\tau$ measures whether at least one lifted proposal reaches 3D IoU $\tau$; proposal-selection accuracy measures whether VPAR selects the target proposal set. Relation subsets may overlap and are therefore diagnostic slices rather than a partition of the overall score.

\paragraph{Implementation and baselines.}
We use Qwen-VL models~\cite{bai2025qwen2} for zero-shot view pruning and VPAR. APG is trained on projected 2D affordance masks, and VPAR is trained with GRPO on APG proposals; 3D IoU $\rho=0.25$ defines a positive proposal and the APG-covered subset. We use $K=25$ views, fusion threshold $\eta=0.3$, GRPO group size 8, and KL coefficient $5\times10^{-3}$. We compare with 3D open-vocabulary methods (OpenMask3D, LERF, and Mask3D-F~\cite{schult2022mask3d}) and VLM-based 2D-to-3D methods (Fun3DU, TASA, and AffordBot). 

For cross-dataset transfer, we use Fun3DU because its released training-free implementation supports unseen scenes.

\subsection{4.2 Quantitative Results}

Table~\ref{tab:main_results} uses the official SceneFun3D evaluator. OpenMask3D and LERF are Fun3DU's split0 reproductions; the other four baselines are reproduced by us and converted to the official format.

\begin{table}[t]
\centering
\renewcommand{\arraystretch}{1.1}
\caption{
\textbf{Official SceneFun3D validation results.}
  Best and second best are \textbf{bold} and \underline{underlined}. \textdagger~marks our reproductions; Mask3D-F is fine-tuned Mask3D~\cite{schult2022mask3d}.
}
\label{tab:main_results}
\setlength{\tabcolsep}{4pt}
\resizebox{\columnwidth}{!}{
\begin{tabular}{lcccc}
\toprule
\textbf{Method} & \textbf{Input} & \textbf{AP} & \textbf{AP$_{50}$} & \textbf{AP$_{25}$} \\
\midrule
\multicolumn{5}{l}{\textit{Non-VLM-based methods:}} \\
\quad OpenMask3D~\cite{takmaz2023openmask3d} & PCD+RGB & -- & 0.00 & 0.00 \\
\quad LERF~\cite{kerr2023lerf} & RGB (multi-view) & -- & 4.90 & 11.30 \\
\quad Mask3D-F$^\dagger$~\cite{schult2022mask3d} & RGB (multi-view) & 0.95 & 1.70 & 2.57 \\
\midrule
\multicolumn{5}{l}{\textit{VLM-based methods:}} \\
\quad TASA$^\dagger$~\cite{he2026task} & RGB+PCD & 0.05 & 0.22 & 1.50 \\
\quad AffordBot$^\dagger$~\cite{wang2026affordbot} & RGB+PCD & \underline{3.90} & \underline{6.37} & 10.70 \\
\quad Fun3DU$^\dagger$~\cite{corsetti2025functionality} & RGB & 1.66 & 4.55 & \underline{12.56} \\
\quad \textbf{ThinkAfford (Ours)} & RGB & \best{4.42} & \best{10.69} & \best{25.46} \\
\bottomrule
\end{tabular}
}
\end{table}

\ours{} reaches 25.46 AP$_{25}$ (+12.90 over Fun3DU) and 10.69 AP$_{50}$ (+4.32 over AffordBot), while its averaged-AP margin is only +0.52. The shrinking lead at stricter thresholds shows that our main gain is finding the correct interaction region; exact high-IoU boundaries remain difficult.

\subsection{4.3 Qualitative Results}
\begin{figure*}[!htbp]
\centering
\IfFileExists{Figures/qualitative_comparison.pdf}{
  \includegraphics[width=0.85\textwidth]{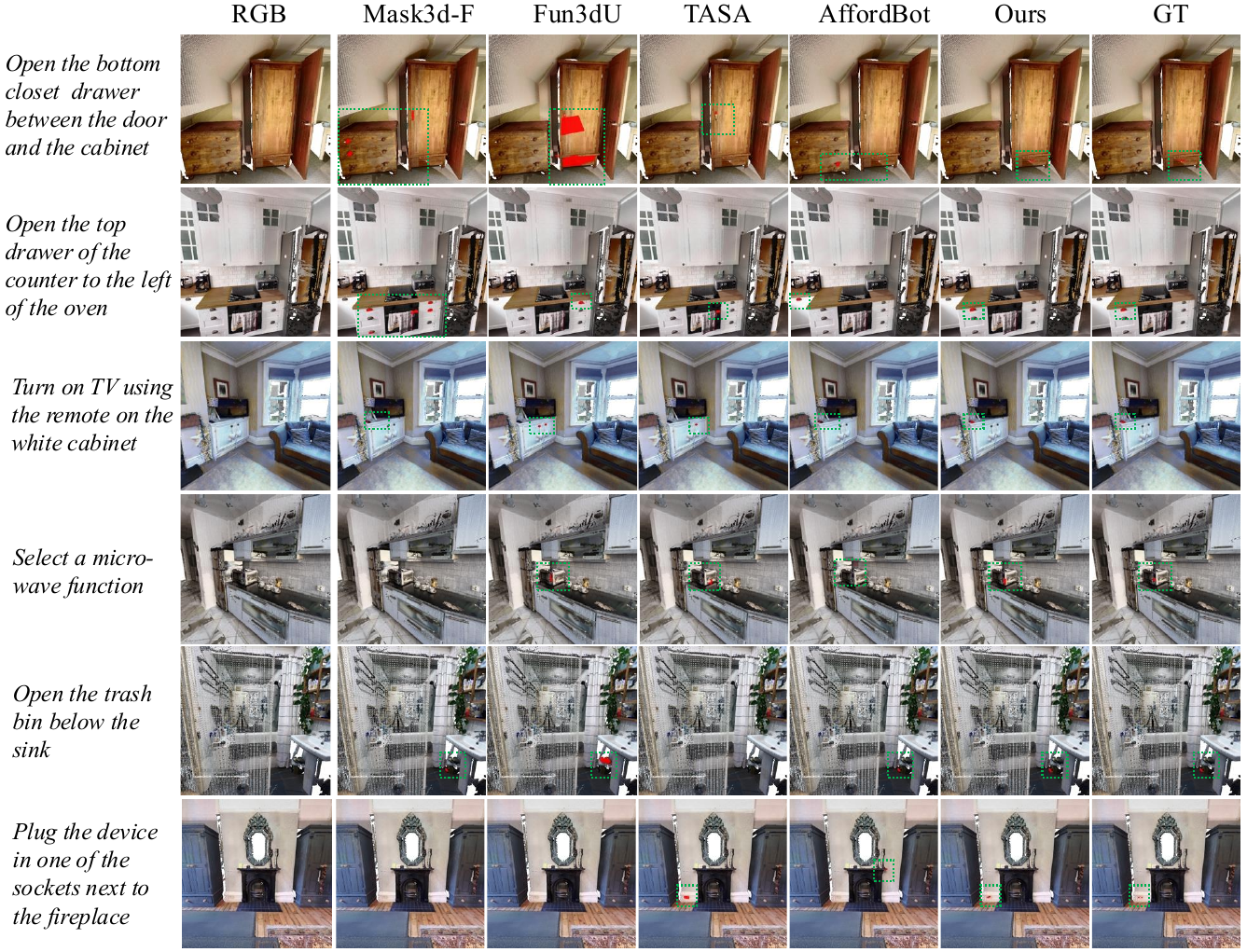}
}{
  \fbox{\begin{minipage}[c][0.22\textheight][c]{0.96\textwidth}
  \centering
  Qualitative comparison placeholder. Replace with
  \texttt{Figures/qualitative\_comparison.pdf} containing columns for
  Mask3D-F, Fun3DU, AffordBot, TASA, \ours{}, and GT.
  \end{minipage}}
}
\caption{
  \textbf{Qualitative comparison.}
  Columns compare four baselines and \ours{} with ground truth on repeated-part disambiguation and functional-region localization.
}
\label{fig:qualitative_comparison}
\end{figure*}

Figure~\ref{fig:qualitative_comparison} reflects the two problems identified in the Introduction. The repeated-drawer examples probe proposal selection: several candidates have the same function, and only ordinal or relative-position constraints identify the target. The remote, microwave, and socket examples probe proposal granularity: the candidate must isolate the actionable region rather than its parent object. Baselines often return the contextual object or wrong repeated instance, whereas \ours{} follows the contact region. These examples are illustrative successes, not evidence that high-IoU boundary errors have disappeared.

\subsection{4.4 Ablation Studies}

\paragraph{Architectural design.}
Table~\ref{tab:ablation} isolates each design choice while retaining the official evaluator, so its deltas can be compared directly with the main result.

\begin{table}[!tbp]
\centering
\caption{
\textbf{Ablation on key components.}
Official AP scores are reported; $\Delta$ is relative to the full model.
}
\label{tab:ablation}
\footnotesize
\setlength{\tabcolsep}{4pt}
\resizebox{\columnwidth}{!}{
\begin{tabular}{lrrrr}
\toprule
Configuration & AP$_{50}$ & $\Delta_{50}$ & AP$_{25}$ & $\Delta_{25}$ \\
\midrule
\textbf{\ours{} full} & \textbf{10.69} & -- & \textbf{25.46} & -- \\
\midrule
\multicolumn{5}{l}{\textit{Proposal generation}} \\
~~w/o APG, parsed-object Grounded-SAM & 6.8 & -3.89 & 18.3 & -7.16 \\
~~APG w/o CoOp prompts & 8.5 & -2.19 & 22.1 & -3.36 \\
~~APG w/o visibility-aware projection & 9.1 & -1.59 & 23.8 & -1.66 \\
\midrule
\multicolumn{5}{l}{\textit{Proposal selection}} \\
~~w/o GRPO (base VLM) & 6.1 & -4.59 & 16.8 & -8.66 \\
~~SFT, think-then-answer & 8.0 & -2.69 & 20.75 & -4.71 \\
~~GRPO w/o format reward & \underline{9.9} & \underline{-0.79} & \underline{24.9} & \underline{-0.56} \\
~~GRPO w/o think format & 8.8 & -1.89 & 22.6 & -2.86 \\
\midrule
\multicolumn{5}{l}{\textit{View selection and fusion}} \\
~~uniform frame sampling & 8.9 & -1.79 & 23.1 & -2.36 \\
~~uniform view weights($w_v=1$) & 9.7 & -0.99 & 24.2 & -1.26 \\
\bottomrule
\end{tabular}
}
\end{table}

The two largest AP$_{25}$ losses come from weakening proposal selection by removing GRPO (-8.66) and weakening proposal generation by replacing APG with parsed-object Grounded-SAM (-7.16). Within APG, affordance prompts matter more than visibility-aware projection (-3.36 vs. -1.66). Within VPAR, the reasoning trace matters more than the format reward (-2.86 vs. -0.56), so the gain is not merely parseable output. Frame selection and learned fusion weights provide smaller but consistent gains (-2.36 and -1.26).

\paragraph{Proposal generation coverage.}
Table~\ref{tab:mechanism_analysis} asks whether proposal generation retains the target region on cropped \texttt{laser\_scan\_5mm} support, using each method's available candidates. 
APG uses an oracle best-view union, so the table diagnoses coverage rather than ranking end-to-end predictions.

\begin{table}[!tbp]
\centering
\caption{
  \textbf{Proposal generation coverage across different affordance types.} Values are R@50/R@25 (\%).
}
\label{tab:mechanism_analysis}
\scriptsize
\setlength{\tabcolsep}{2.0pt}
\renewcommand{\arraystretch}{1.05}
\resizebox{\columnwidth}{!}{
\begin{tabular}{@{}lrrrrr@{}}
\toprule
Affordance (\#Queries) & Mask3D-F & TASA & AffordBot & Fun3DU & \textbf{APG} \\
\midrule
foot\_push (1) & 0.0/0.0 & 0.0/0.0 & \textbf{100.0/100.0} & 0.0/0.0 & 0.0/0.0 \\
hook\_pull (136) & 2.2/2.9 & 1.5/20.6 & 11.8/16.2
& \underline{13.2}/\underline{45.6} & \textbf{41.2/71.3} \\
hook\_turn (56) & 16.1/23.2 & 1.8/14.3 & \underline{32.1}/35.7
& 12.5/\underline{37.5} & \textbf{48.2/85.7} \\
key\_press (23) & 4.3/4.3 & 13.0/30.4 & 8.7/21.7
& \underline{43.5}/\underline{82.6} & \textbf{78.3/91.3} \\
pinch\_pull (110) & 2.7/5.5 & 16.4/45.5 & 13.6/24.5
& \underline{29.1}/\underline{70.0} & \textbf{61.8/90.0} \\
plug\_in (24) & 0.0/0.0 & 0.0/8.3 & 0.0/\underline{20.8}
& \underline{4.2}/16.7 & \textbf{41.7/79.2} \\
rotate (40) & 0.0/0.0 & 5.0/10.0 & 0.0/7.5
& \underline{25.0}/\underline{47.5} & \textbf{52.5/62.5} \\
tip\_push (42) & 2.4/2.4 & 7.1/14.3 & 0.0/4.8
& \underline{16.7}/\underline{26.2} & \textbf{42.9/64.3} \\
unplug (13) & 0.0/0.0 & 7.7/30.8 & \underline{15.4}/30.8
& \underline{15.4}/\underline{46.2} & \textbf{53.8/69.2} \\
\midrule
\textbf{Overall (445)} & 3.8/5.6 & 6.7/24.5 & 12.1/20.0
& \underline{19.6}/\underline{49.2} & \textbf{50.6/77.5} \\
\bottomrule
\end{tabular}}
\end{table}

\begin{figure*}[!htbp]
\centering
\includegraphics[width=0.9\linewidth]{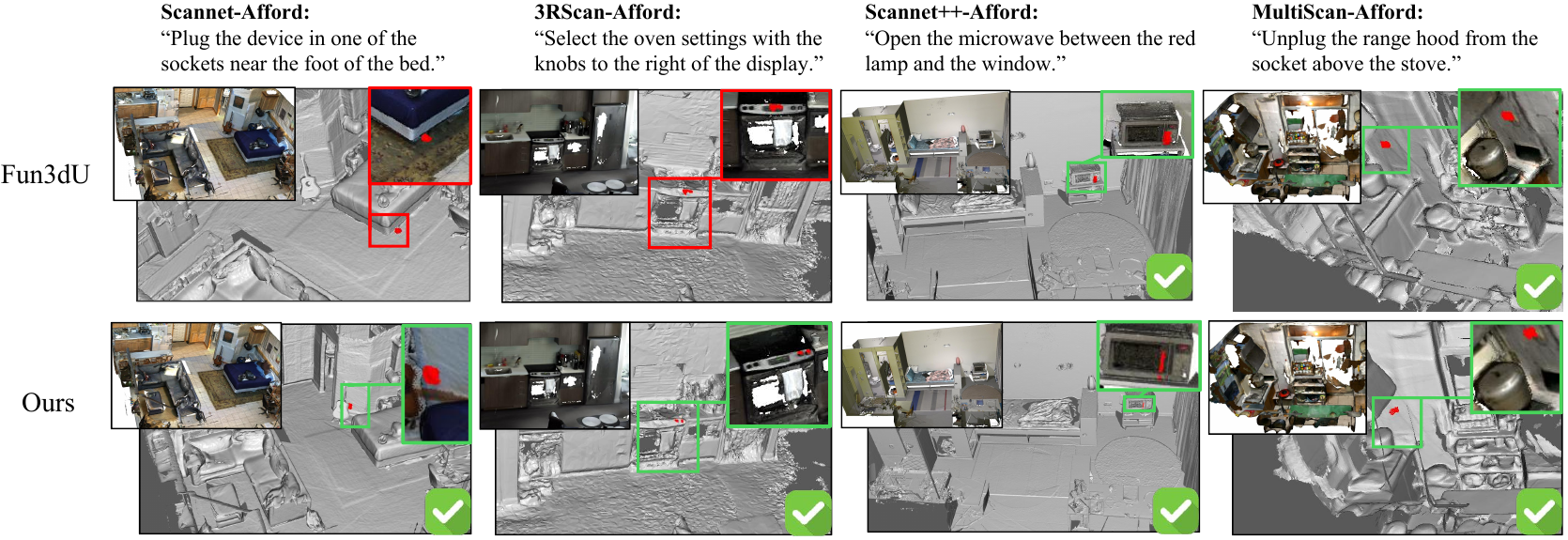}
\caption{
  \textbf{Qualitative zero-shot cross-dataset transfer.}
  Predictions on the four unseen datasets in Table~\ref{tab:cross_dataset}; boxes magnify the red 3D affordance regions.
}
\label{fig:cross_dataset_qualitative}
\end{figure*}

\begin{figure*}[!htbp]
\centering
\includegraphics[width=\textwidth]{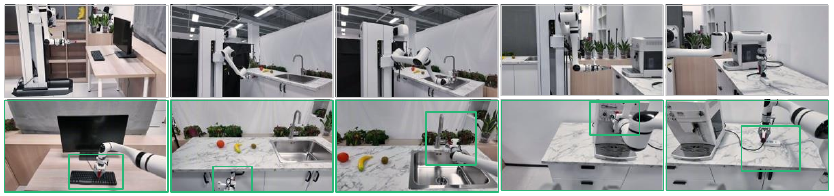}
\caption{
  \textbf{Qualitative real-world deployment.}
  Five manipulation examples with scene-level robot approach views (top) and close-up execution views of the localized interaction regions (bottom).
}
\label{fig:real_robot_deployment}
\end{figure*}

APG reaches 77.5 R@25 and 50.6 R@50, exceeding Fun3DU by 28.3/31.0 points. On hook\_pull, the largest class, gains remain 25.7/28.0, so rare classes do not drive the average. The sole reversal is the one-sample foot\_push class. APG's 26.9-point R@25--R@50 gap reveals the remaining issue: targets are often present but not tightly bounded.

\paragraph{Proposal selection accuracy.}
Table~\ref{tab:vpar_reasoning} evaluates whether VPAR selects the correct proposal set once APG has covered the target. Accuracy is measured on the APG-covered subset; relation slices overlap.

\begin{table}[!tbp]
\centering
\caption{
  \textbf{Proposal selection accuracy.}
  Accuracies are percentages and relation subsets may overlap; AP$_{50}$/AP$_{25}$ is end-to-end on all queries.
}
\vspace{-0.2cm}
\label{tab:vpar_reasoning}
\scriptsize
\setlength{\tabcolsep}{2.5pt}
\renewcommand{\arraystretch}{1.05}

\begin{tabular*}{\columnwidth}{@{\extracolsep{\fill}}lrrrrr@{}}
\toprule
\multirow{2}{*}{Setting}
  & \multicolumn{4}{c}{Selection Acc. (\%) $\uparrow$}
  & \multicolumn{1}{c}{End-to-end $\uparrow$} \\
\cmidrule(lr){2-5}\cmidrule(l){6-6}
  & Overall & Spatial & Ordinal & Part-of & AP$_{50}$/AP$_{25}$ \\
\midrule
\multicolumn{6}{l}{\textit{Qwen-VL2.5}} \\
~~3B Base  & 19.6 & 19.2 & 11.5 & 17.6 & 3.08/9.13 \\
~~3B +GRPO & \underline{43.6} & \underline{46.0} & \underline{46.9}
  & \underline{46.8} & \underline{6.23/16.90} \\
~~7B Base  & 40.9 & 42.9 & 43.1 & 41.2 & 5.88/15.03 \\
~~7B +GRPO & \textbf{54.2} & \textbf{57.1} & \textbf{60.8}
  & \textbf{61.6} & \textbf{8.08/23.90} \\
\midrule
\multicolumn{6}{l}{\textit{Qwen-VL3-8B}} \\
~~Base (no tuning) & 51.3 & 54.1 & 41.2 & 60.3 & 6.1/16.8 \\
~~SFT & 63.4 & 67.2 & 55.8 & 71.4 & 8.0/20.75 \\
~~GRPO w/o format & \underline{68.1} & \underline{72.3} & \underline{64.5}
  & \underline{75.8} & \underline{9.9/24.9} \\
~~GRPO w/o trace & 65.8 & 68.9 & 59.7 & 73.2 & 8.8/22.6 \\
~~\textbf{Full GRPO (Ours)} & \textbf{72.1} & \textbf{76.5} & \textbf{69.8}
  & \textbf{79.3} & \textbf{10.69/25.46} \\
\bottomrule
\end{tabular*}
\vspace{-0.5cm}
\end{table}

Full GRPO raises Qwen-VL3-8B selection from 51.3\% to 72.1\% (+8.7 over SFT), with the largest gains on ordinal (+28.6) and spatial (+22.4) queries. This is not scale alone: 3B+GRPO exceeds untuned 7B in both selection (43.6 vs. 40.9) and AP$_{25}$ (16.90 vs. 15.03). Removing the trace costs 6.3 selection points and 2.86 AP$_{25}$, versus 4.0 and 0.56 without the format reward; reasoning contributes more downstream than compliance. Proposal coverage and selection accuracy use different query sets and are not numerically composable, but together they expose the remaining errors: APG misses 22.5\% of all queries at IoU 0.25, while VPAR misselects 27.9\% of covered queries. Improving proposal tightness and relation-sensitive selection therefore remains complementary rather than interchangeable.

\subsection{4.5 Cross-Dataset Generalization}

We compare zero-shot transfer with Fun3DU under the same point-mask protocol. \ours{} is trained only on SceneFun3D, Fun3DU is training-free, and neither method uses target-dataset samples.

\begin{table}[!tbp]
\centering
\caption{
  \textbf{Zero-shot cross-dataset transfer.}
  AP$_{50}$/AP$_{25}$ (\%); neither method uses target-dataset samples.
}
\vspace{-0.2cm}
\label{tab:cross_dataset}
\scriptsize
\setlength{\tabcolsep}{2.0pt}
\renewcommand{\arraystretch}{1.08}
\begin{tabular*}{\columnwidth}{@{\extracolsep{\fill}}lccccc@{}}
\toprule
Method & 3RScan & MultiScan & ScanNet & ScanNet++ & Average \\
\midrule
Fun3DU
  & \underline{12.1}/\underline{22.2}
  & \textbf{29.7}/\underline{49.7}
  & \underline{15.0}/\underline{25.0}
  & \underline{15.5}/\underline{24.0}
  & \underline{18.1}/\underline{30.2} \\
\textbf{\ours{}}
  & \textbf{20.0/32.3}
  & \underline{26.4}/\textbf{57.2}
  & \textbf{21.6/28.3}
  & \textbf{23.8/29.2}
  & \textbf{23.0/36.8} \\
\bottomrule
\end{tabular*}
\end{table}

\ours{} raises average AP$_{50}$/AP$_{25}$ from 18.1/30.2 to 23.0/36.8, winning AP$_{25}$ on all datasets and AP$_{50}$ on three. MultiScan is the informative exception: AP$_{25}$ rises by 7.5 while AP$_{50}$ falls by 3.3. Thus the transferred prior finds the approximate region, but boundary quality does not adapt equally across reconstruction domains, matching the proposal-tightness gap in Table~\ref{tab:mechanism_analysis}.

\subsection{4.6 Real-World Deployment}
We pass \ours{}'s scene-coordinate contact region to an existing robot execution stack. Figure~\ref{fig:real_robot_deployment} presents five real-robot manipulation episodes and checks sensing-to-coordinate transfer outside the benchmark. This is qualitative, not a manipulation success-rate study: \ours{} predicts neither force, orientation, nor trajectory, and we do not evaluate the controller.

\section{5 Conclusion}
We present \ours{}, a task-driven 3D affordance grounding framework that decomposes the problem into high-recall affordance proposal generation and instruction-grounded reasoning. 
This decomposition separates two key failure modes in existing approaches: missing the interaction region during proposal construction and confusing similar candidates during language-guided selection. 
APG generates fine-grained affordance-aware proposals without predefined object or part prompts, while VPAR selects proposals under the full instruction and is optimized with GRPO using 3D grounding rewards. 
Experiments on SceneFun3D demonstrate the effectiveness of \ours{}, with both proposal coverage and reasoning optimization contributing to improved grounding performance.
Cross-dataset and real-world results further indicate transfer beyond the benchmark. 

\bibliography{main}

\end{document}